\documentclass[letterpaper, 10 pt, conference]{ieeeconf}
\IEEEoverridecommandlockouts
\usepackage{graphics} 
\usepackage{wrapfig} 
\usepackage[linesnumbered,ruled]{algorithm2e}
\graphicspath{{img/}}
\usepackage{epsfig} 
\usepackage{amsmath} 
\usepackage{amssymb}  
\usepackage{amsthm}
\usepackage{bigints}
\usepackage{pifont}
\def\-{\raisebox{.75pt}{-}}
\usepackage{mathtools}
\usepackage{cite}
\usepackage[makeroom]{cancel}
\usepackage[table]{xcolor} 
\usepackage[center]{subfigure}
\usepackage{multirow}
\usepackage{hyperref}
\usepackage{booktabs}
\usepackage{subfigure}
\usepackage{epstopdf}
\usepackage{textcomp}
\usepackage{makecell}
\usepackage{bm}
\usepackage{caption}
\usepackage{algpseudocode,algorithm} 
\usepackage{color, soul}
\usepackage{mathptmx}
\usepackage[11pt]{moresize}
\usepackage{hhline}
\definecolor{ForestGreen}{RGB}{34,139,34}

\newcommand\copyrighttextfinal{%
	\scriptsize\copyright\ 2022 IEEE. Personal use of this material is permitted. Permission from IEEE must be obtained for all other uses, in any current or future media, including reprinting/republishing this material for advertising or promotional purposes, creating new collective works, for resale or redistribution to servers or lists, or reuse of any copyrighted component of this work in other works.}%
\newcommand\copyrightnotice{%
	\begin{tikzpicture}[remember picture,overlay]%
	\node[anchor=south,yshift=10pt] at (current page.south) {{\parbox{\dimexpr\textwidth-\fboxsep-\fboxrule\relax}{\copyrighttextfinal}}};%
	\end{tikzpicture}%
}
\usepackage{tikz}
\usepackage{textcomp}
\def\BibTeX{{\rm B\kern-.05em{\sc i\kern-.025em b}\kern-.08em
    T\kern-.1667em\lower.7ex\hbox{E}\kern-.125emX}}

\usepackage{eso-pic}

\newcommand{\subparagraph}{}
\definecolor{amaranth}{rgb}{0.9, 0.17, 0.31}
\definecolor{bleudefrance}{rgb}{0.19, 0.55, 0.91}

\usepackage{hyperref}
\hypersetup{
pdftitle={DeepFusion: A Robust and Modular 3D Object Detector for Lidars, Cameras and Radars},
pdfsubject={IEEE/RSJ International Conference on Intelligent Robots and Systems (IROS 2022)},
pdfauthor={Florian Drews, Di Feng, Florian Faion, Lars Rosenbaum, Michael Ulrich and Claudius Gl\"aser},
pdfkeywords={}
}

\begin{document}
\bstctlcite{IEEEexample:BSTcontrol}
\title{\Huge DeepFusion: A Robust and Modular 3D Object Detector for Lidars, Cameras and Radars
	\thanks{$^*$All authors are with Robert Bosch GmbH, Corporate Research, 71272 Renningen, Germany,  \url{florian.drews@bosch.com}}
}
\author{Florian Drews, Di Feng, Florian Faion, Lars Rosenbaum, Michael Ulrich and Claudius Gl\"aser$^{*}$}
\maketitle
\copyrightnotice
\begin{abstract}
We propose DeepFusion, a modular multi-modal architecture to fuse lidars, cameras and radars in different combinations for 3D object detection. Specialized feature extractors take advantage of each modality and can be exchanged easily, making the approach simple and flexible. Extracted features are transformed into bird's-eye-view as a common representation for fusion. Spatial and semantic alignment is performed prior to fusing modalities in the feature space. Finally, a detection head exploits rich multi-modal features for improved 3D detection performance. 
Experimental results for lidar-camera, lidar-camera-radar and camera-radar fusion show the flexibility and effectiveness of our fusion approach.
In the process, we study the largely unexplored task of faraway car detection up to 225~meters, showing the benefits of our lidar-camera fusion. Furthermore, we investigate the required density of lidar points for 3D object detection and illustrate implications at the example of robustness against adverse weather conditions. Moreover, ablation studies on our camera-radar fusion highlight the importance of accurate depth estimation.
\end{abstract}

\section{Introduction}\label{sec:introduction}
Safe and accurate 3D object detection is a core technology in autonomous driving. All subsequent components like tracking, prediction and planning heavily depend on the detection performance~\cite{wong2020testing}. Errors in perception of other traffic participants can potentially propagate through the system, leading to severe failure of the autonomous vehicle. To prevent such errors, the perception system needs careful design, which remains a challenging research question~\cite{janai2017computer}. 

Multiple sensors and different modalities, mainly lidars, RGB-cameras and radars, are often used to approach this object detection challenge~\cite{nuscenes2019}. Multiple sensors improve system redundancy, and different modalities increase detection robustness, as their complementary physical properties can be exploited to overcome different driving scenarios, where a single-modality fails. For example, lidars and cameras suffer from strong degradation in foggy weather conditions~\cite{bijelic2020seeing}, whereas radars stay relatively unaffected. On the other hand, radars and monocular cameras suffer from sparse or imprecise depth estimation~\cite{ma2021delving,wang2019pseudo,reading2021categorical}, which can be compensated by the dense and accurate lidar point cloud.

Over the last years, great progress has been made in the task of 3D object detection with modalities of lidars~\cite{yin2021center,yang2018pixor,shi2020pv}, cameras~\cite{huang2021bevdet,wang2019pseudo} and radars~\cite{ulrich2022improved,ulrich2021deepreflecs,niederlohner2022self,svenningsson2021radar}.
This trend is fuelled by public large-scale multi-modal datasets like nuScenes~\cite{nuscenes2019} and Waymo~Open~Dataset~\cite{sun2019scalability}. However, the research community mainly focuses on close-range 3D object detection up to $75$ meters. We emphasize the importance of far-range object detection, as early detection of other traffic participants and their actions could enable safer, faster and smoother overall system reactions, especially on highway scenarios. Therefore, the target of this paper is an object detector design that is scalable above $200$ meters range.

A major challenge of working with multiple sensors and modalities is the fusion of this multitude of redundant and complementary sensor data. At the interface between multiple sensors and the output of perception, the fusion has an important impact on the overall system performance. Most existing works focus on fusing lidar and RGB-camera sensors for 3D object detection~\cite{chen2017multi,qi2017frustum,ku2017joint,liang2018deep,xu2021fusionpainting, yin2021center,vora2020pointpainting,yoo20203d,feng2020leveraging}. 
Less explored are camera-radar fusion~\cite{chadwick2019distant,nabati2021centerfusion,kim2020grif} and lidar-camera-radar fusion~\cite{hendy2020fishing,ravindran2022camera}, the latter are on the tasks of semantic heatmap prediction and 2D object detection respectively. 
We argue the necessity of developing a simple and flexible lidar-camera-radar fusion network for 3D object detection, because radar sensors are complementary to lidars and cameras regarding their measuring principle, and radar technology is developing quickly with increasing market demands~\cite{yang2020radarnet,kim2020grif,meyer2019automotive}.

In this work, we propose a modular network architecture to fuse lidars, cameras and radars for accurate, robust, and long-range 3D object detection. The approach employs exchangeable feature-extractors to yield well-optimized architectures for single-modal detectors. Extracted rich features of each modality are then transformed into a common bird's-eye-view representation for convenient fusion in the shared latent space. This network design enables us to easily investigate the fusion of different combinations of modalities, with a focus on lidar-camera, lidar-camera-radar and camera-radar fusions. 

Our main contributions are:
\begin{itemize}
\item We propose a novel modular architecture to fuse lidar, camera and radar for 3D object detection in a simple and effective manner. 
\item In-depth investigation and discussion on different combinations of modalities.
\item Study faraway 3D object detection of up to $225$ meters whereas popular benchmarks like nuScenes\cite{nuscenes2019} and Waymo\cite{sun2019scalability} focus on rather close range of $50$ and $75$ meters respectively.
\end{itemize}

\section{Background and Related Works}
\label{sec:related_works}
In this section, we briefly discuss the properties of cameras, radars, and lidars in 3D perception, and summarize deep learning methods for multi-modal object detection. We refer interested readers to our survey paper~\cite{feng2019modal} for a more comprehensive overview.

\subsection{Cameras, Radars, and Lidars}
RGB camera images capture detailed texture information of objects, and are widely applied for object classification. However, cameras do not directly provide depth information, making the 3D vision task challenging, especially in the mono-camera setting~\cite{ma2021delving,wang2019pseudo,reading2021categorical}. Radar points give us azimuth speed, and radial distance, which are helpful to locate dynamic objects. Radars are also robust against various lighting and weather conditions~\cite{yang2020radarnet}. However, 3D object detection using radars is limited by low resolution and erroneous elevation estimates~\cite{yang2020radarnet, ulrich2021deepreflecs}. Lidar points provide accurate depth information of the surroundings, and come with a higher resolution of object details in 3D space compared to camera images or radar points with erroneous depth estimation or sparsity~\cite{hendy2020fishing} respectively. Therefore, many 3D object detection leader-boards (such as KITTI~\cite{Geiger2012CVPR} or nuScenes~\cite{nuscenes2019}) are driven by lidar-based solutions~\cite{shi2020pv,shi2020part,yin2021center}.

\subsection{Multi-modal Object Detection}
RGB cameras and lidars are the most common sensors for fusion in the literature~\cite{chen2017multi,qi2017frustum,ku2017joint,liang2018deep,xu2021fusionpainting, yin2021center,vora2020pointpainting,yoo20203d,feng2020leveraging}. In addition, \cite{liu2016bmvc} combines RGB images with thermal images, \cite{liang2019multi} combines lidar point clouds with HD maps, and recently an increasing number of works combine RGB camera images with radar points~\cite{chadwick2019distant,nabati2021centerfusion,kim2020grif}. Only~\cite{hendy2020fishing} proposes a general framework to fuse lidars, radars, and RGB camera images. 

State-of-the-art fusion networks follow either the two-stage~(\cite{ku2017joint,liang2019multi,xu2021fusionpainting,yin2021center,chen2017multi,qi2017frustum,feng2020leveraging}) or the one-stage object detection pipeline~(\cite{chadwick2019distant, vora2020pointpainting,liang2018deep,wang2017fusing}). For example, MV3D~\cite{chen2017multi} proposes a two-stage object detector with RGB camera images and lidar points. In the first stage, camera images and lidar points are processed by sensor-specific networks to extract high-level features. The lidar branch also generates region proposals in order to crop lidar and camera feature maps. In the second stage, the cropped features are fused by a small detection head. AVOD~\cite{ku2017joint} extends MV3D by fusing features both in the first and the second stages. 3D-CVF~\cite{yoo20203d} applies a gating mechanism to learn the weights of each modality. MMF~\cite{liang2019multi} adds additional tasks such as image depth completion and ground estimation to the fusion pipeline. Frustum-PointNet~\cite{qi2017frustum} employs a pre-trained image detector to generate 2D object proposals and frustums in the 3D space, and applies a PointNet~\cite{qi2017pointnet} to do object detection using lidar points within the frustums. In the one-stage pipeline, sensors can be fused at one specific layer~\cite{chadwick2019distant, vora2020pointpainting}, or at multiple layers~\cite{liang2018deep,wang2017fusing}. For example, PointPainting~\cite{vora2020pointpainting} applies a pre-trained semantic segmentation network to predict pixel-wise image semantics, and appends the semantic scores to the corresponding lidar points for further 3D object detection. ContFuse~\cite{liang2018deep} gradually fuses the feature maps from the camera and the lidar branches via continuous fusion layers.

Our proposed fusion architecture can be categorized into the one-stage object detection pipeline: each sensor modality is processed by a separate branch. The sensor-specific features are projected onto the bird's eye view, aligned, and fused at a fusion layer. 

\section{DeepFusion Architecture} \label{sec:methodology}
\begin{figure*}[!tpb]
	\centering
	\includegraphics[width=0.9\textwidth]{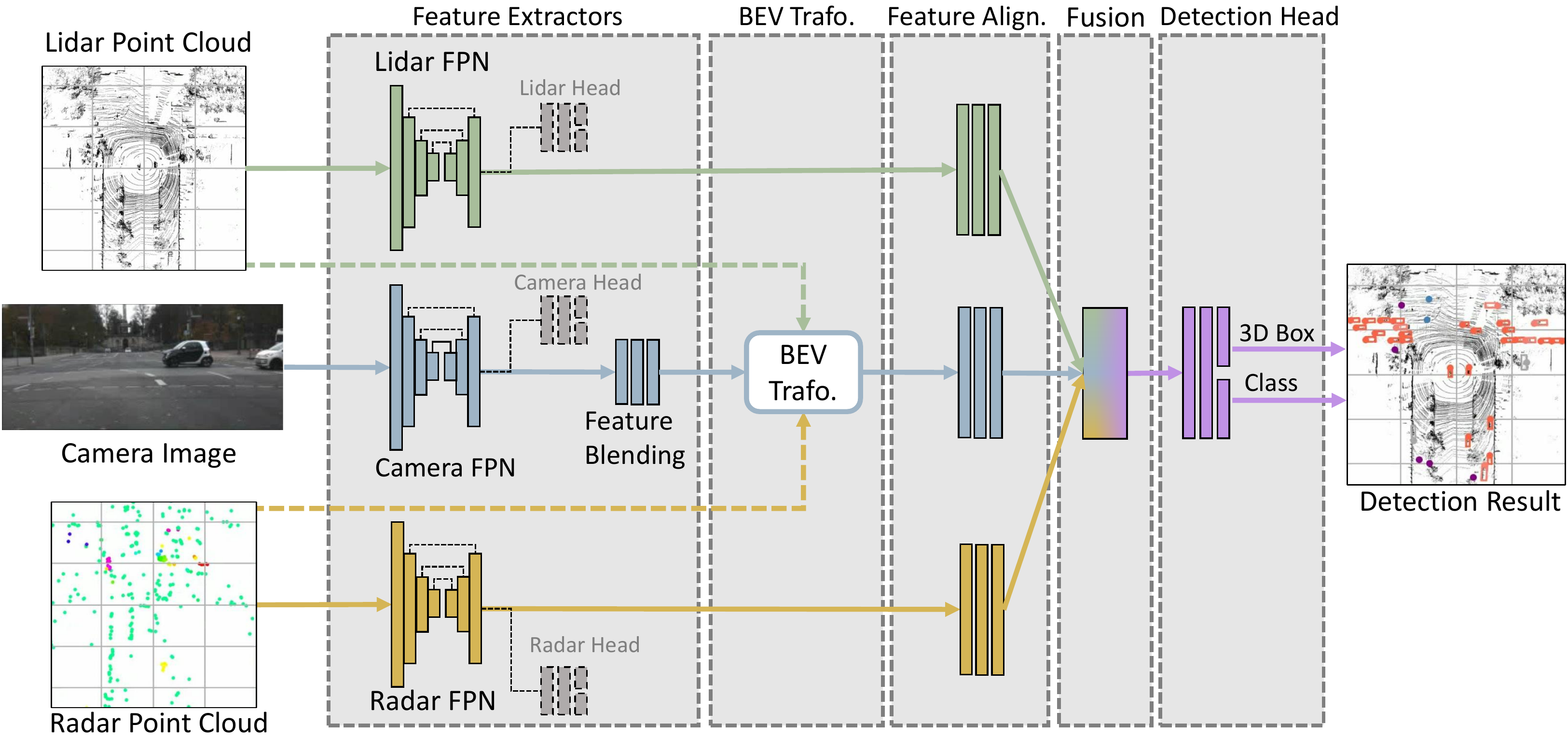}
	\caption{The proposed DeepFusion architecture for modular feature fusion of lidar, camera and radar in BEV. A separate feature pyramid network (FPN) for each modality is employed to extract modality-specific features, which are then transformed into common BEV representation. Features are aligned and aggregated before a detection head provides the classification and regression output.} 
	\label{fig:architecture}
\end{figure*}

Our modular and flexible architecture design shown in Figure~\ref{fig:architecture} builds upon strong feature-extractors for rich encodings of single-modality input data. The birds-eye-view (BEV) transformation module maps these features into a common representation space. In the BEV representation, the objects' sizes are well-preserved with small variances and occlusions, making it well-suited for 3D object detection~\cite{chen2017multi}. Next, the feature alignment module aligns the latent representations between modalities, before being aggregated in the fusion module. In the end, a detection head operating on fused features provides the classification and regression outputs for 3D object detection.

\subsection{Feature Extractors}
For each modality (lidar, camera and radar), there is a separate object detector available to extract features from the respective input data. Well-studied 2D-CNN architectures~\cite{yang2018pixor,he2017mask} are used for this purpose, as their dense 2D outputs are rich in detail. Nevertheless, using other popular approaches like 3D-sparse-convolutions~\cite{shi2020pv} are conceivable as well. The object detectors consist of a feature-pyramid-network (FPN)~\cite{lin2017feature} extracting multi-scale feature maps, and a detection head for classification and box regression outputs. The FPN serves as feature extractor for fusion, whereas the detection head is used for pre-training and as an axillary loss in an end-to-end training setup. 

\subsection{Camera FPN}
A camera sensor provides RGB images $I$ of shape $(H, W, 3)$ with height $H$ and width $W$ as input to the camera FPN. First, the multi-scale feature maps are extracted by the FPN~\cite{lin2017feature}. Afterwards, those feature maps are linearly upscaled to a common scale $Z$, concatenated and processed by several convolution layers for multi-scale feature blending~\cite{lin2017feature}. The outputs are high-quality features $F^C$ of shape $(Z H, Z W, K)$, with $K$ being the number of channels.

\subsection{Lidar/Radar FPN}
Following PIXOR~\cite{yang2018pixor}, the lidar and radar point clouds are represented by the occupancy grid maps on the bird's-eye-view (BEV) plane for feature extraction, with a grid size of height $X$ and width $Y$. A FPN processes the input grid with 2D convolutions by downscaling and subsequentially upscaling this representation to extract features. The outputs are feature maps $\{F^M_{bev},M \in {(L, R)}\}$ of the shape $(S X, S Y, K_{bev})$ with scaling factor $S$, number of channels $K_{bev}$ and modality $M$ for lidar~($L$) and radar~($R$) respectively. These BEV feature maps serve as latent representation for the respective branch, and get processed in the feature alignment module later on. 

\subsection{BEV Transformation}
Latent representations from feature extractors are transformed into the bird's-eye-view for a common spatial representation. Lidar and radar features are already represented in the BEV space, whereas camera features require an image-to-BEV transformation. There are different approaches in the literature for such a transformation, which can be categorized based on the input representation. Camera-only approaches like OFT~\cite{roddick2018orthographic} and ``Lift, splat, shoot"~\cite{philion2020lift} cast-out an image into 3D by estimating depths, and pooling vertical pillars into a BEV representation. The results are dense-depths, but spatially imprecise due to inaccurate depth estimation from mono-camera~\cite{ma2021delving,wang2019pseudo,reading2021categorical}. Point-cloud approaches~\cite{wang2018fusing,liang2018deep,liang2019multi,vora2020pointpainting} require a 3D point-cloud, preferably from a lidar sensor, to guide the transformation. With known sensor calibration between lidar and camera, each point is projected onto the camera image and the BEV grid, building an association of features between the image pixels and BEV cells. The result is a sparsely occupied representation due to the sparsity of the point cloud, but spatially precise thanks to the depth accuracy of lidar measurements. 

DeepFusion uses the point-cloud driven approach from~\cite{wang2018fusing}, as spatial accuracy is important for 3D object detection. Furthermore, any sensor modality can be used to provide the point cloud for image transformation: lidar or radar points are directly used for transformation, and for camera images, the object centroid predictions from the camera detector are employed as the sparse pseudo-points.
During fusion, the point cloud from lidars, radars, and cameras can be aggregated, making the fusion robust against failure from one type of sensor. We use the mean pooling to merge multiple camera features, if they are projected onto the same BEV grid cell. As a result, the image feature map $F^C$ of shape $(Z H, Z W, K)$ is transformed to the BEV plane unfolding the feature map $F^{C}_{bev}$ with shape $(S X, S Y, K_{bev})$.

\subsection{Feature Alignment}
Inputs to the feature alignment module are the densely-occupied BEV feature maps $F^{L}_{bev}$ and $F^{R}_{bev}$ from the lidar and radar FPNs, and the sparsely-occupied transformed feature map $F^{C}_{bev}$ from camera, thus these are diverse spatial representations. Additionally, these feature maps originate from different modalities and FPN backbones, and encode different semantic representations. In this regard, a feature alignment module, built up from several convolution layers, is separately applied on $\{F^{M}_{bev},M \in{(L,C,R)}\}$, in order to align the respective representations spatially and semantically. 
The outputs are feature maps $\{F^{M}_{align},M \in{(L,C,R)}\}$ of the same shape $(S X, S Y, K_{bev})$.

\subsection{Fusion}
The fusion module receives $\{F^{M}_{align}, M \in{(L,C,R)}\}$ as the aligned feature-maps from the lidar, the camera, and the radar branches, respectively. The fusion module has the task to combine these different modalities in the latent space. To get the fused feature-map $F=\rho(\{F^{M}_{align},M \in{(L,C,R)}\})$, a fusion operation $\rho$ is applied, which can be a fixed operation like pooling or weighted averaging~\cite{feng2019modal}, or a learnable operation like attention~\cite{vaswani2017attention}. The output of the module is a fused feature-map $F$ of shape $(S X, S Y, K_{bev})$ and scale $S$.

\subsection{Detection Head}
The detection head receives the fused feature-map $F$ to generate classification and regression outputs for 3D bounding boxes. Due to the rich multi-modal features encoded in the feature-map $F$, we find that a small head with a few convolution layers is enough for generating more precise and robust 3D objects than those from a single-modal detector.

\section{Implementation Details} \label{sec:implementation}
\subsubsection{Feature Extractors}
For the camera branch, we use MaskRCNN~\cite{he2017mask} pre-trained on COCO~\cite{lin2014microsoft} and feature maps with downscaling factor~${\{\frac{1}{4}, \frac{1}{8}, \frac{1}{16}, \frac{1}{32}\}}$ and $256$~channels. The feature blending model consists of five 3x3-conv-layers with $96$~channels, relu activation except for the last and 1x1-conv for first layer.
We employ the simple, yet effective lidar 3D object detector, PIXOR~\cite{yang2018pixor}, for the lidar branch operating on a BEV occupancy grid with cell size of $0.1$ meters, height $X=140$ meters in driving-direction and width $Y=80$ meters. Feature maps with scales $S \in{(\frac{1}{4}, \frac{1}{2})}$ and channels $K_{bev}=96$ are later on used for detecting cars and pedestrians respectively.
For the radar branch, we use the same approach as for lidar, except for a few adaptions to account for radar characteristics following~\cite{ulrich2022improved}. Radar point-clouds have a lower resolution compared to lidar point-clouds, and provide additional attributes like velocity $v$ and radar-cross-section $rcs$. Therefore, Pillar~Feature~Net~\cite{niederlohner2022self,lang2018pointpillars} is used to encode the radar features for a BEV grid with cell size of $0.5$~m. The radar feature maps are upscaled to match the lidar BEV resolution. 

\subsubsection{Feature Alignment and Fusion}
The feature alignment model follows the same architecture as the feature blending model. For fusion we apply the additive fusion operation
$F=F^{L}_{align}+F^{C}_{align}+F^{R}_{align}$, which has been found to perform well in our experiments.

\subsubsection{Training Setup} 
For the lidar, camera and radar FPNs, we use the weights from individual detectors pre-trained for 3D object detection. The fusion detection head is initialized by the pre-trained weights from the lidar or radar detector. When training the fusion model, we freeze the FPNs, and train all other parts of the architecture, including the feature alignment module, the fusion layer, and the fusion detection head. An Adam optimizer~\cite{kingma2014adam} is used, and the training curve converges after $10$ epochs. We find this simple training strategy already achieves promising results, and requires less training time compared to the end-to-end training strategy.

We use the same training loss for sensor-specific detectors and the fusion network. For classification we use the focal loss~\cite{lin2018focal}, denoted by $L_{cls}$, and for bounding box regression we use the $L_2$ loss, denoted by $L_{reg}$. The final training loss $L$ is a weighted sum of classification and regression:
\begin{equation}
L = w_{cls} L_{cls} + w_{reg} L_{reg}.
\end{equation}

\section{Experimental Results} \label{sec:experimental_result}
We conduct detailed experiments to study the properties of each sensor modality and to verify the proposed fusion architecture. Section~\ref{subsec:comparison} compares the detection performance among sensor specific detectors and different fusion combinations, which is followed by some qualitative results in Section~\ref{subsec:qualitative_results}. Section~\ref{subsec:ablation_study} shows the ablation studies mainly for LC, LCR and CR fusion schemes, regarding the nice/bad weather conditions, the number of points, the detection distance, and the detection performance for faraway objects up to $225$ meters. Finally, Section~\ref{subsec:nuscenes_benchmark} benchmarks our fusion network on the challenging nuScenes dataset~\cite{nuscenes2019}. 

\subsection{Experimental Setup}\label{subsec:experimental_setup}
The experiments are mainly conducted on our internal multi-modal Bosch dataset, with recordings from several lidars, cameras, and radars. The data was recorded in several countries of Europe (major cities, rural areas, and highways) and in different weather conditions (sunny, rainy, cloudy). Objects were categorized into 17 classes, and labelled with 3D bounding boxes. This work uses a subset of the Bosch dataset, with approx. 10k training frames (the training set) and 3k validation frames (the validation set). Additionally, we use the public nuScenes dataset~\cite{nuscenes2019} to benchmark our method with other state-of-the-art object detectors. The dataset was captured in Boston and Singapore with readings from lidars, cameras, and radars. We show the results of training on 28K frames and evaluation on 6k validation frames.

Following the nuScenes object detection benchmark~\cite{nuscenes2019}, the standard Average Precision (AP) metric is used to evaluate 3D detections. An AP score ranges within $[0\%, 100\%]$, with a larger value indicating better detection performance. We report AP with different localization thresholds and its mean (mAP) over all thresholds. A localization threshold is defined by the euclidean distance of bounding box centroids between a detection and its ground truth ($0.5$, $1.0$, $2.0$, $4.0$ meters). All detections are evaluated within the $140$ meters range, or mentioned otherwise. 

We evaluate the detection performance on the car or pedestrian objects, the two major object classes in autonomous driving research~\cite{Geiger2012CVPR,nuscenes2019}. For the ablation studies we use a loose localization threshold ($4.0$ meters), in order to compensate the effect of temporal misalignments in different sensors, especially in the long range and for dynamic objects.

\subsection{A General Comparison of Detection Performance}\label{subsec:comparison}
We compare the performance among several sensor-specific detectors (C, R, L) and the fusion networks with different sensor combinations (CR, LC, LCR). Table~\ref{tab:bosch_detection_results} illustrates the AP scores of the car class at different location thresholds, as well as their mean scores (mAP). The table shows a clear AP improvement from sensor fusion compared to sensor-specific networks. CR outperforms C and R by more than $20\%$ and $10\%$ mAP, respectively, and LC improves L by $5\%$ mAP. The best detection performance is achieved by LCR, which fuses all sensor modalities. These experimental results verify the effectiveness of our proposed fusion architecture.

\begin{figure} [tpb]
	\centering
	\includegraphics[width=0.7\linewidth]{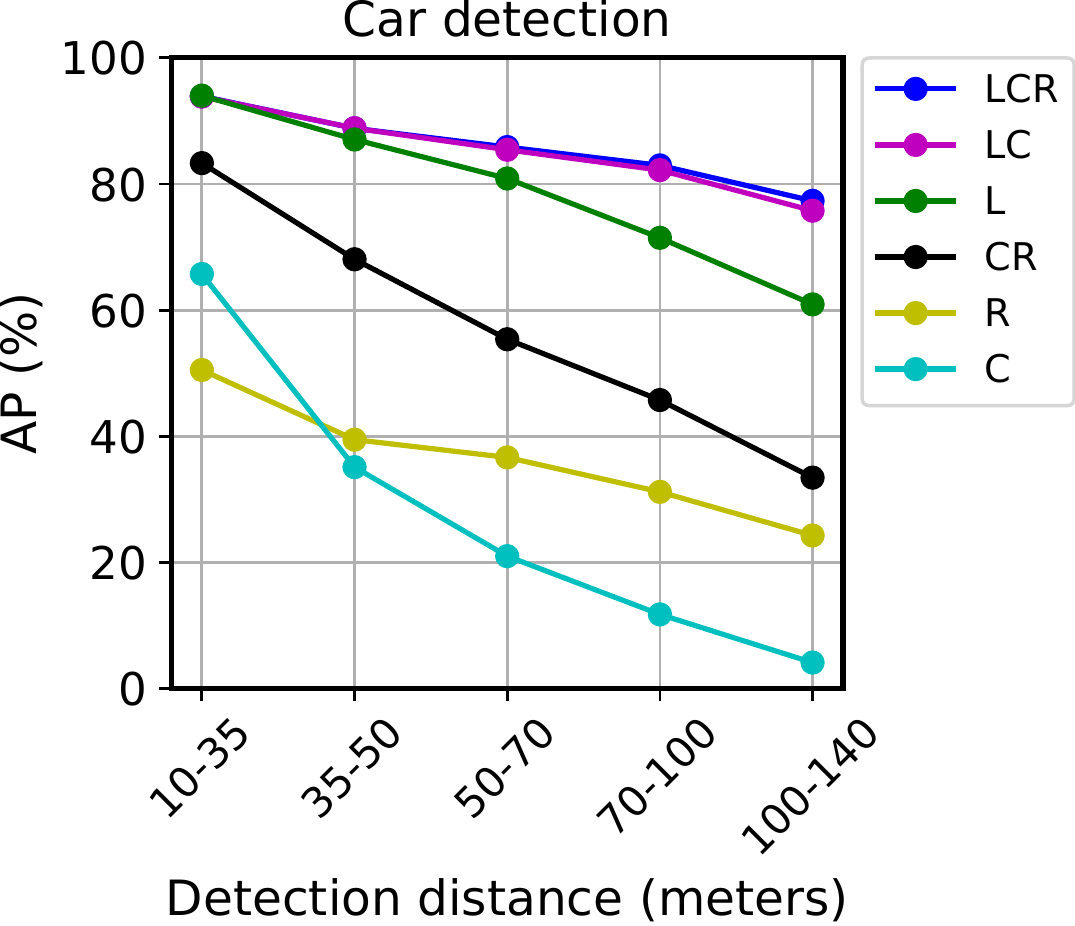}
	\caption{A comparison of detection performance in AP@4.0 over distance. Results for car class on Bosch validation frames. C: camera, R: radar, L: lidar.} \label{fig:modalities_AP}
\end{figure}

To study the properties of each sensor modality, Figure~\ref{fig:modalities_AP} shows the evolutions of AP scores at $4.0$ meters localization threshold with respect to the detection range. The camera-only detector C outperforms its radar counterpart R in the near range ($10-35$ meters), but degrades drastically at larger distance, indicating the large depth estimation errors in vision. The detectors with lidar points (L, LC, LCR) perform better than those without lidar points (C, R, CR) with a big margin. For example, L surpasses CR by $20\%$ AP at a distance of $100-140$ meters, showing the importance of using lidar points for long-range detection. Furthermore, LC increasingly improves L at larger distance up to nearly $20\%$ AP. We hypothesize this is because the object texture features provided by cameras images are helpful for detection, especially when lidar points become sparse in the long range. Finally, we observe a marginal improvement of $1\%$ AP when comparing LCR with LC, indicating that radars encode some object features complementary to those from cameras and lidars, such as velocities. 

\begin{table}[tpb]
	\begin{tabular}[h]{c|c c c c | c}
		\hline
		Modality & AP@0.5 & AP@1.0 & AP@2.0 & AP@4.0 & mAP\\
		\hline
		C & 1.1 & 4.9 & 14.4 & 27.6 & 12.0 \\
		R & 3.6 & 15.3 & 30.4 & 36.4 & 21.5 \\
		CR & 6.2 & 22.9 & 45.4 & 57.2 & 32.9 \\ \hline
		L & 58.8 & 73.2 & 78.4 & 78.9 & 72.3 \\
		LC & \textbf{61.3} & 77.7 & 84.3 & 85.2 & 77.1 \\	
		LCR & 61.1 & \textbf{78.3} & \textbf{84.9} & \textbf{85.8} & \textbf{77.5} \\
		\hline
	\end{tabular}
	\caption{\label{tab:bosch_detection_results} A comparison of detection performance for the car class on the Bosch validation frames. C: camera, R: radar, L: lidar.}
\end{table}

\subsection{Qualitative Results}\label{subsec:qualitative_results}
\begin{figure}[!tpb]
	\centering
	\includegraphics[width=0.99\linewidth]{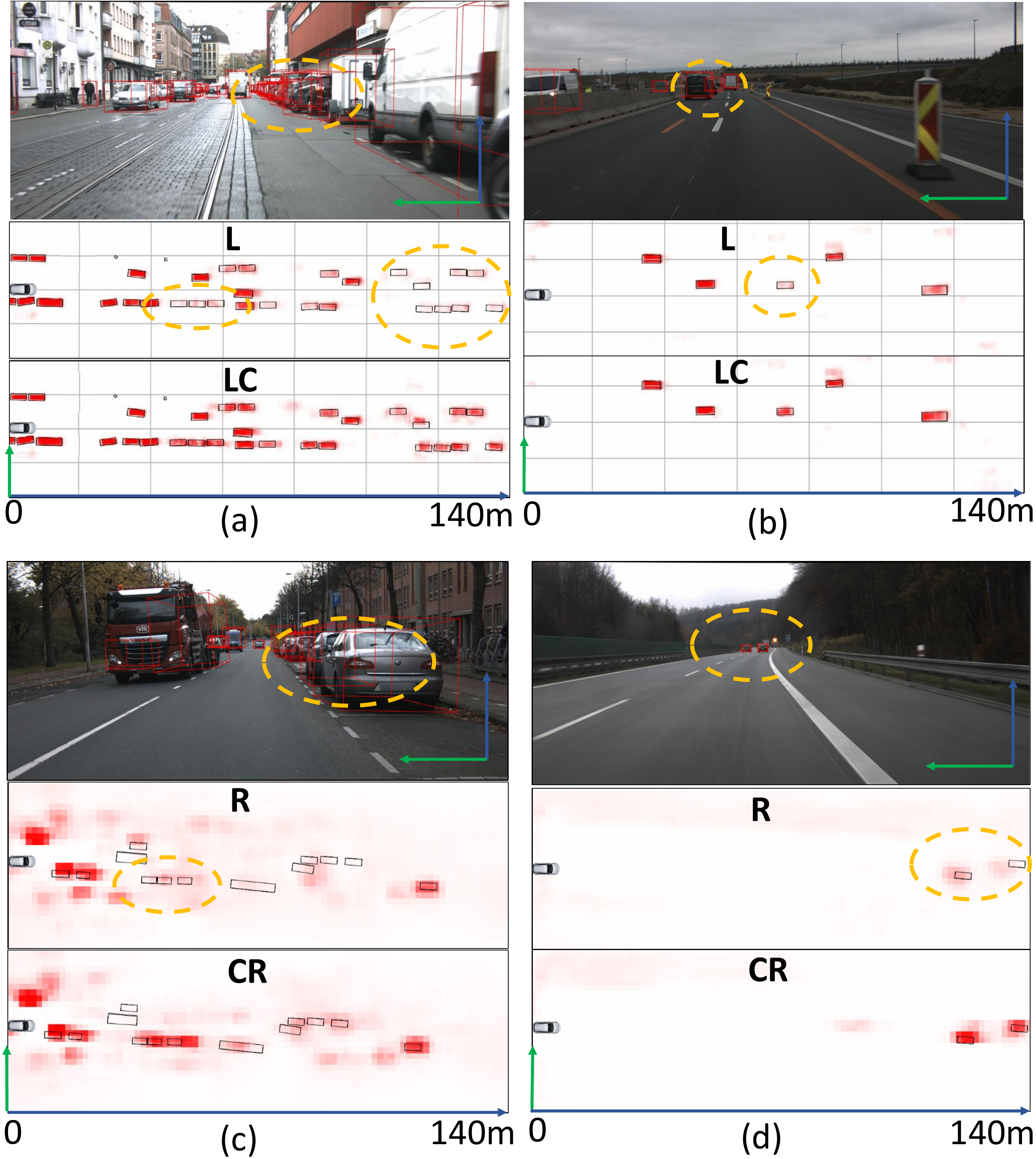}
	\caption{Qualitative results for car detection on the Bosch dataset. BEV heatmaps visualize the classification confidence of the detector in red. Lidar-camera fusion~(LC) results at the top row with camera image, lidar heatmap and fusion heatmap. Radar-camera fusion~(CR) at the bottom row with camera image, radar heatmap and fusion heatmap. Missed detections of baseline model marked in orange.}
	\label{fig:examples}
\end{figure}

Qualitative results of our LC and CR fusion models for car detection on the Bosch dataset are visualized in Figure~\ref{fig:examples}. We compare fusion and the corresponding single-modal baseline model, by visualizing objectness heatmaps before non-maximum-suppression. The red color indicates a high confidence of car objects. The fusion models are able to reliably detect objects missed by the baseline models highlighted in orange. Figure~\ref{fig:examples}(a) and Figure~\ref{fig:examples}(b) show LC results with camera image, the heatmap of baseline L and the heatmap of LC. In (a) LC is able to detect parked cars with high level of occlusion, which is missing by the L model. In (b) LC detects a highly occluded car on the highway, which shows low objectness score in the lidar-only model. Figure~\ref{fig:examples}(c) and Figure~\ref{fig:examples}(d) show CR fusion results with camera images,  the heatmap of baseline R and the heatmap of the CR model. In (c) parked cars were not detected, because partly-occluded static objects are particularly difficult for radar, whereas fusion with camera is able to detect them. In (d) radar struggles for faraway objects, which are detected reliably by the CR fusion.

\subsection{Detailed Analysis and Ablation Study}\label{subsec:ablation_study}

\subsubsection{LCR Fusion} \label{subsubsec:lcr_fusion}
We study the effect of weather conditions on our LCR fusion. For this purpose the validation set of the Bosch dataset is split into the nice weather set (sunny or cloudy days with dry roads) and the bad weather set (rainy days with wet roads). The lidar-based models L, LC and LCR are evaluated on these splits. Figure~\ref{fig:weather_LCR_abs_Car} shows the AP scores for the car class with $4.0$ meters threshold. Each model shows a performance drop at the bad weather conditions, most distinctive at increasing distances. The largest performance gap is observed in model L, with $11\%$ AP drop in the range of $100-140$ meters. For further quantification of this nice-bad weather gap, we use the mRAPD metric as a measure for robustness against bad weather conditions. It is calculated by the mean value of difference AP in bad weather $AP^{(a)}$ relative to that in nice weather $AP^{(b)}$ averaged over $D$ distance bins:
\begin{equation}\label{eq:rel_ap_decrease}
\text{mRAPD} = \frac{1}{D}\sum_{d\in{D}}{\frac{AP^{(a)}_d - AP^{(b)}_d}{ AP^{(b)}_d}}.
\end{equation}

\begin{figure}[!htpb]
	\centering
	\begin{minipage}{1\linewidth}
		\centering
		\subfigure[]{\label{fig:LCR_ablation_study_num_points}\includegraphics[width=0.48\textwidth]{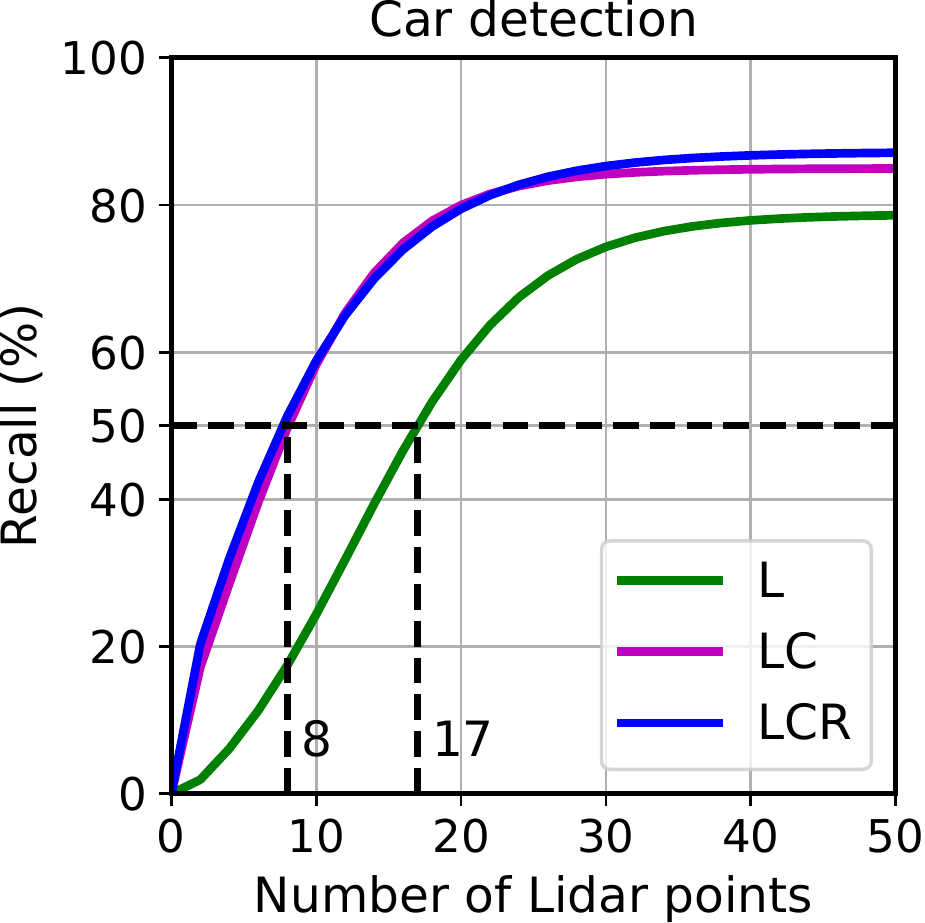}}
		\subfigure[]{\label{fig:LCR_ablation_study_distance}\includegraphics[width=0.48\textwidth]{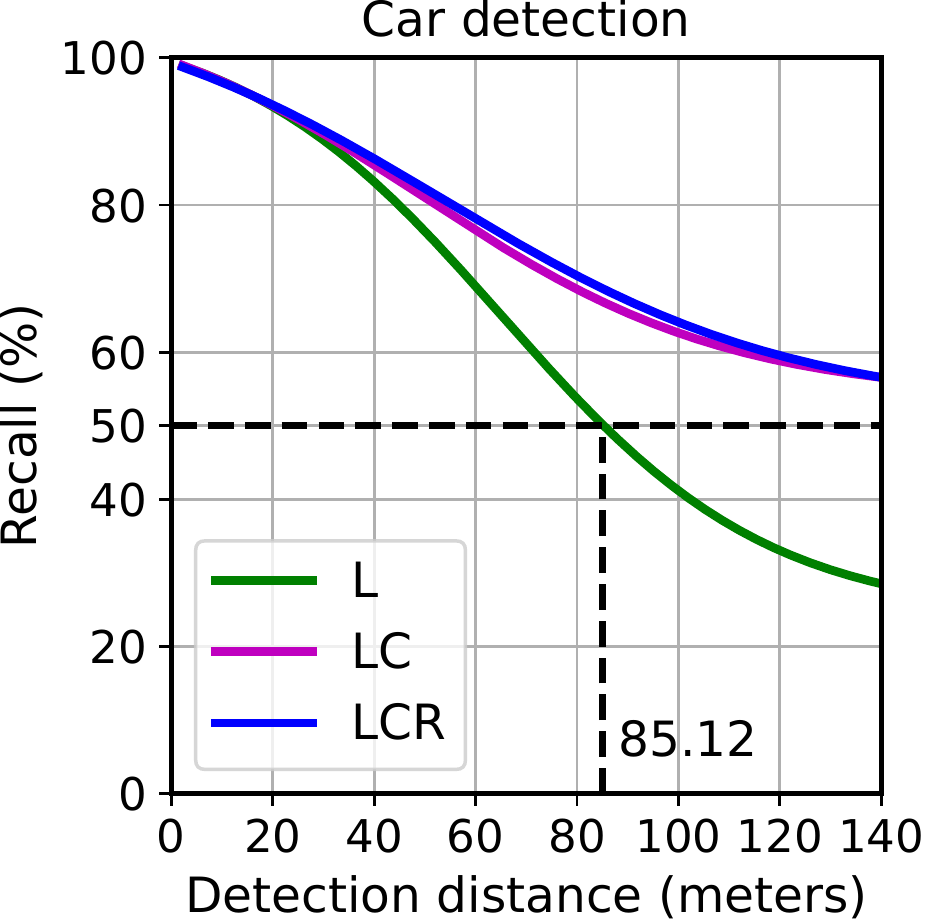}}
		\caption{The recall rates of car detection with respect to the number of lidar points within a bounding box (a) and the detection range up to $140$ meters (b).}
	\end{minipage}
\end{figure}

Using the mRAPD metric, we find that the lidar-only model L is mostly affected by bad weather among the models with $-7.1\%$ mRAPD. This is because rainy weather deteriorates the point cloud quality and reduces the number of points per object, causing the ``missing point problem"~\cite{xu2021spg}. Since there are no complementary sensors, the model L can not compensate the information loss of reducing lidar points. In this case, fusion with camera images remedies the ``missing point problem", resulting in $-4.0\%$ mRAPD for the LC model. The LCR model is most robust against weather conditions with $-2.7\%$ mRAPD, because radars are less affected by rainy weather than lidars or cameras. 

We further study the performance of LCR fusion with respect to the density of lidar points and the detection distance. The models L, LC and LCR are compared. Figure~\ref{fig:LCR_ablation_study_num_points} and Figure~\ref{fig:LCR_ablation_study_distance} show the recall rates of car detection with respect to the number of lidar points within a bounding box and the detection range up to $140$ meters, respectively. Note that in Figure~\ref{fig:LCR_ablation_study_num_points} we only plot up to $50$ points, as the recall rates are found to be converged with more points. Fusing lidar points with camera images significantly improves the recall rates compared to the lidar-only network. While the model L achieves a recall rate above $50\%$ using more than $17$ lidar observations or within $85$ meter detection range, the model LC requires only $8$ lidar points, and keeps the recall rate to nearly $60\%$ even at $140$ meters. LCR slightly improves LC by $1\%$ recall rate at the same lidar density or detection distance.

\subsubsection{CR Fusion}
When performing camera-radar fusion (CR), the 3D locations from radar observations (R points) and 3D object centroids predicted by the camera-branch network (C points) are used to extract camera features. Those camera features are reprojected onto the BEV plane, in order to do fusion with radar features. Therefore, the fusion performance is highly dependent on the ``quality'' of 3D points. This ablation study evaluates the impact of the origin of 3D points used for the CR fusion.

\begin{figure}[!tpb]
	\centering
	\begin{minipage}{1\linewidth}
		\centering
		\subfigure[]{\label{fig:weather_LCR_abs_Car}\includegraphics[width=0.48\textwidth]{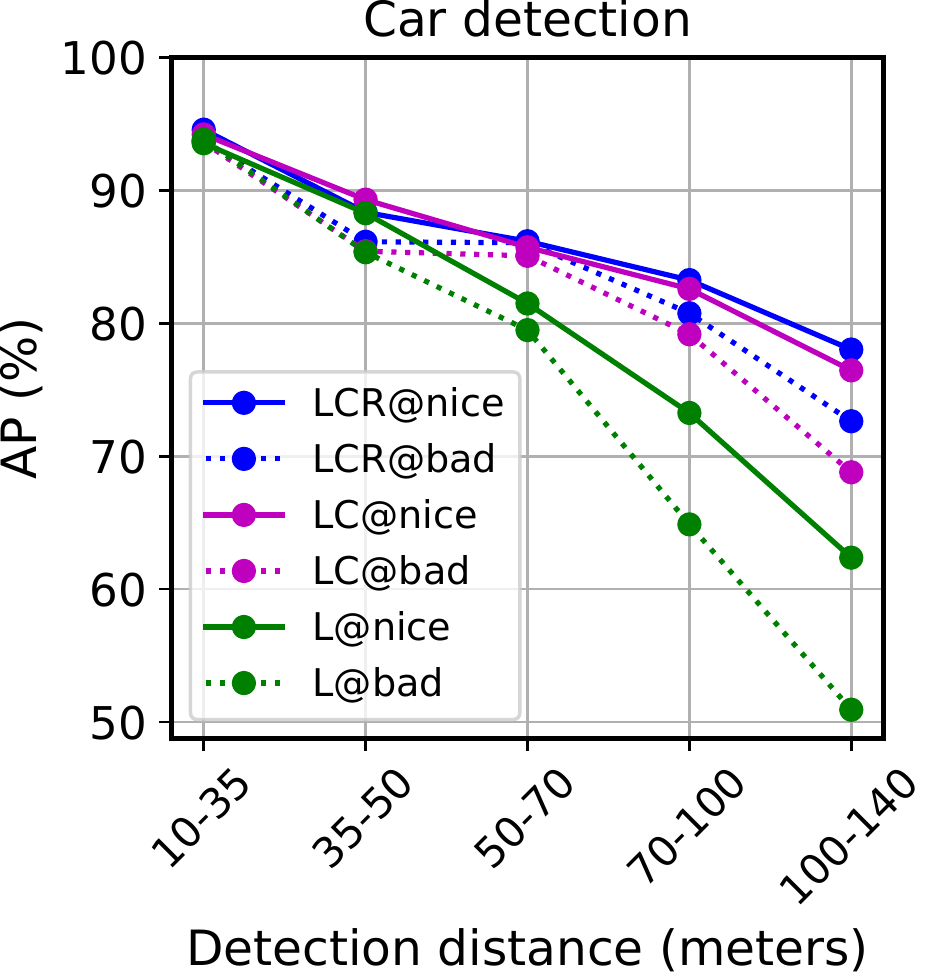}}
		\subfigure[]{\label{fig:CR_ablation_study}\includegraphics[width=0.48\textwidth]{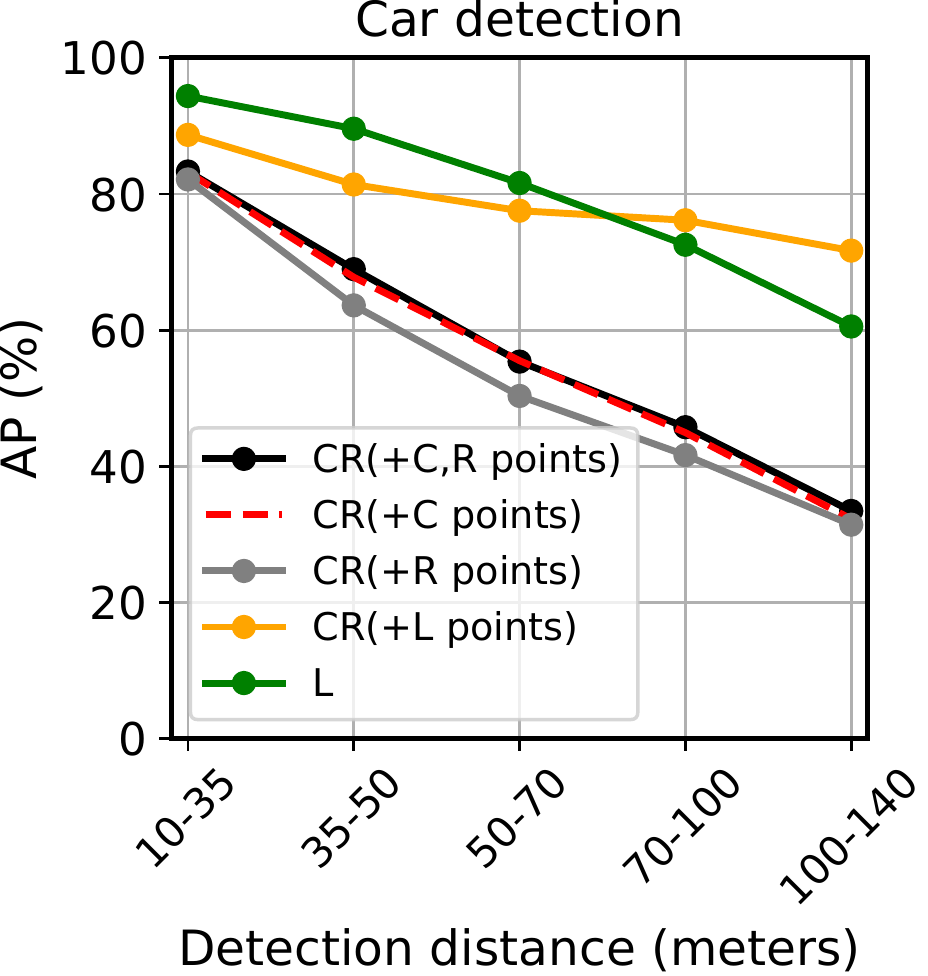}}
		\caption{(a) Detection results in AP@4.0m of L, LC and LCR models with respect to nice and bad weather conditions. (b) Detection performance of CR fusion using different methods to extract camera features.}
	\end{minipage}
\end{figure}

We alternate types of points to project camera features onto the BEV grid during inference, based on a CR model trained both with R points and C points (C,R points). We compare the inference using C and R points ``CR(+C,R points)'' with those either use C points ``CR(+C points)'' or R points ``CR(+R points)''. In addition, we test the performance of CR fusion when using lidar point locations (L points) to extract camera features, called ``CR(+L points)''. It serves as a upper-bound of CR fusion, assuming that C and R points could match the density and accuracy of L points. Finally, we use the lidar-only detector ``L'' as a benchmark. The results are demonstrated in Figure~\ref{fig:CR_ablation_study}.

From the figure we can see the importance of C points in the CR fusion: ``CR(+C points)'' outperforms ``CR(+R points)'' by up to $5\%$ AP, and adding R points in addition to C points does not improve fusion, as the similar performance between ``CR(+C points)" and ``CR(+C,R points)" illustrates. One reason might be that C points could capture more object locations with useful camera features, whereas R points are inaccurate in locations or missing due to measurement errors and occlusion. Furthermore, comparing ``CR(+L points)" with other CR models, we observe a significant performance gain by up to $40\%$ AP. ``CR(+L points)" even outperforms the lidar-only detector at a distance larger than $70$ meters, even though it was trained without any lidar points. The result indicates that accurate and dense 3D locations are important when extracting camera features. 

\subsubsection{Faraway Object Detection}\label{subsubsec:faraway_od}
Table~\ref{tab:faraway_objects} compares the L and LC models for faraway car detection. Though both models are trained only up to $140$ meters range, the proposed fusion architecture constantly improves the lidar-only detector beyond the training range, and achieves nearly $30\%$ AP above $200$ meters. The results verify the model's scalability in the long range.

\begin{table}[!tpb]
	\centering
	\begin{tabular}[h]{c|c|c}
		\hline
		Modality & AP($140-200$m) & AP($200-225$m)\\
		\hline
		L & 48.3 & 16.4\\
		LC & 64.3 \bf{(+16.0)} & 29.5 \bf{(+13.1)}\\
		\hline
	\end{tabular}
	\caption{\label{tab:faraway_objects} AP for the challenging task of faraway car detection of up to $225$ meters on the Bosch dataset. Models were trained on data up to $140$ meters and evaluated beyond, showing generalization capabilities. }
\end{table}

\subsection{Evaluation on the nuScenes Dataset}\label{subsec:nuscenes_benchmark}
We evaluate our lidar-based models on the challenging nuScenes benchmark~\cite{nuscenes2019} for 3D object detection. As the MaskRCNN network trained on COCO~\cite{lin2014microsoft} is not allowed in the nuScenes detection leaderboard by their pre-training rules, we use the lightweight EfficientNetB0~\cite{tan2019efficientnet} architecture with pre-trained weights from ImageNet as our camera backbone.
Furthermore, we downscale the input images to $576 \times 256$ from the native resolution of $1600 \times 900$ before feeding them into the model. We find this resolution sufficient to achieve good fusion results with reduced inference time. As lidar and radar points are very sparse in the dataset, we follow the common practice for nuScenes~\cite{nuscenes2019} to aggregate up to $10$ lidar scans and up to $7$ radar scans with ego-motion compensation for our models. The aggregated lidar scans are used in the BEV transformation with the current camera image. Only camera images of keyframes are used.

We compare fusion methods also with respect to their lidar baseline performance. For this purpose the nuScenes validation split is used, as respective papers~\cite{xu2021fusionpainting,yoo20203d} report their lidar baseline performance on it. 

Table~\ref{tab:nuscenes_detection_results} compares the AP scores for car and pedestrian classes on the nuScenes validation set. Our focus is on these two most represented classes to rule out effects of the pronounced class imbalance problem investigated in~\cite{zhu2019class}. 
With an AP of $77.9$ and $77.1$ for car and pedestrian detection, our PointPillar-alike lidar model (L) provides a solid baseline. Fusing lidar points with camera images (LC model) improves the lidar model (L) by $+2.5$\% AP and $+6.6$\% AP scores for car and pedestrian classes, respectively, showing the importance of the texture feature from camera images, especially for detecting small objects.
The best results are achieved by the lidar-camera-radar model (LCR) with $+3.7\%$ and $+7.5\%$ AP gains over the lidar model (L), verifying the effectiveness of our fusion design. By replacing our lidar model (L) with an even stronger lidar backbone, such as CenterPoint~\cite{yin2021center}, we expect higher AP scores on fusion.

\begin{table}[!tpb]
\begin{tabular}[h]{c|c|c|c}
	\hline
	Method & Modality & AP(Car) & AP(Pedestrian) \\
	\hline
	SECOND\cite{yan2018second,yoo20203d} & L & 69.2 & 58.6 \\
	InfoFocus\cite{wang2020infofocus} & L & 77.6 & 61.7 \\
	3D-CVF\cite{yoo20203d} & L & 78.2 & 68.7 \\
	PointPillars\cite{lang2019pointpillars,xu2021fusionpainting} & L & 80.7 & 70.8 \\
	CenterPoint\cite{yin2021center,xu2021fusionpainting} & L & \bf{84.7} & \bf{83.4} \\
	DeepFusion (Ours) & L & 77.9 & 77.1 \\
	\hline
	3D-CVF\cite{yoo20203d} & LC & 79.7 (+1.5) & 71.3 (+2.6) \\
	FusionPainting*\cite{xu2021fusionpainting} & LC & \bf{83.5 (+2.8)} & 82.9 (+12.1) \\
	DeepFusion (Ours) & LC & 80.4 (+2.5) & 83.7 (+6.6) \\
	DeepFusion (Ours) & LCR & 81.6 (+3.7) & \bf{84.6 (+7.5)} \\
	\hline
\end{tabular}
\caption{
	The detection performance on the nuScenes benchmark~\cite{nuscenes2019}. We show AP (\%) for the car and pedestrian classes on the validation set. The AP gains of fusion method over their corresponding lidar baselines are shown in brackets. *For FusionPainting we compare with the variant based on PointPillars.}
\label{tab:nuscenes_detection_results}
\end{table}

\section{Discussion and Conclusion} \label{sec:conclusion}
We have presented DeepFusion, a modular fusion architecture for accurate long-range 3D object detection using lidars (L), cameras (C), and radars (R). The method first extracts modality-specific features using separate backbones. Those features are then aligned, transformed, and fused in a common latent space based on the bird's eye view (BEV) representation. A small detection head processes the fused features for final prediction. Through this simple and flexible fusion design, single-modal feature-extractors can be easily exchanged to keep up with the progress in their respective fields, and different sensor modalities and fusion strategies can be easily combined. In this study, we focus on LC, LCR and CR fusions. Extensive experiments highlight the importance of sensor fusion in increasing detection range up to $225$ meters, and in improving robustness against adverse weather conditions on rainy days and wet roads. In the future, we intend to introduce more approaches for BEV representation into our fusion architecture, such as Lift-Splat-Shoot~\cite{philion2020lift} and BEVDet~\cite{huang2021bevdet}.

\bibliographystyle{IEEEtran}
\bibliography{bibliography}

\end{document}